\documentclass[11pt]{article}

\usepackage[final]{acl}

\usepackage{times}
\usepackage{latexsym}

\usepackage[T1]{fontenc}

\usepackage[utf8]{inputenc}

\usepackage{microtype}

\usepackage{inconsolata}

\usepackage{graphicx}

\usepackage{amsmath}
\usepackage{amssymb}

\usepackage{booktabs}
\usepackage{multirow}
\usepackage{tabularx}

\usepackage{algorithm}
\usepackage{algorithmic}

\usepackage{caption}
\usepackage{enumitem}
\usepackage{subcaption}
\usepackage{tcolorbox}
\usepackage[table,x11names,dvipsnames]{xcolor}
\usepackage{hyperref}
\definecolor{dartgreen}{HTML}{00693e}
\definecolor{refcolor}{HTML}{9F363A}

\hypersetup{
    colorlinks=true,
    linkcolor=Bittersweet,
    citecolor=dartgreen,
    filecolor=magenta,      
    urlcolor=dartgreen,
    }
    
\title{Addressing Overthinking in Large Vision-Language Models via Gated Perception-Reasoning Optimization}

\author{
\textbf{Xingjian Diao}$^1$,
\textbf{Zheyuan Liu}$^2$,
\textbf{Chunhui Zhang}$^1$,
\textbf{Weiyi Wu}$^1$,
\textbf{Keyi Kong}$^1$,
   \\
\textbf{Lin Shi$^3$,}
\textbf{Kaize Ding$^4$,}
\textbf{Soroush Vosoughi$^1$,}
\textbf{Jiang Gui$^1$}
\\
 $^1$Dartmouth College,  $^2$University of Notre Dame,  $^3$Cornell University,  $^4$Northwestern University
 \\
   \texttt{xingjian.diao.gr@dartmouth.edu}
}
\begin{document}
\maketitle

\begin{abstract}
Large Vision-Language Models (LVLMs) have exhibited strong reasoning capabilities through chain-of-thought mechanisms that generate step-by-step rationales. However, such slow-thinking approaches often lead to overthinking, where models produce excessively verbose responses even for simple queries, resulting in test-time inefficiency and even degraded accuracy. Prior work has attempted to mitigate this issue via adaptive reasoning strategies, but these methods largely overlook a fundamental bottleneck: visual perception failures. We argue that stable reasoning critically depends on low-level visual grounding, and that reasoning errors often originate from imperfect perception rather than insufficient deliberation. To address this limitation, we propose Gated Perception-Reasoning Optimization (GPRO), a meta-reasoning controller that dynamically routes computation among three decision paths at each generation step: a lightweight fast path, a slow perception path for re-examining visual inputs, and a slow reasoning path for internal self-reflection. To learn this distinction, we derive large-scale failure attribution supervision from approximately 790k samples, using teacher models to distinguish perceptual hallucinations from reasoning errors. We then train the controller with multi-objective reinforcement learning to optimize the trade-off between task accuracy and computational cost under uncertainty. Experiments on five benchmarks demonstrate that GPRO substantially improves both accuracy and efficiency, outperforming recent slow-thinking methods while generating significantly shorter responses.
\end{abstract}

\begin{figure}[tb]
\centering
\resizebox{0.48\textwidth}{!}{
\includegraphics{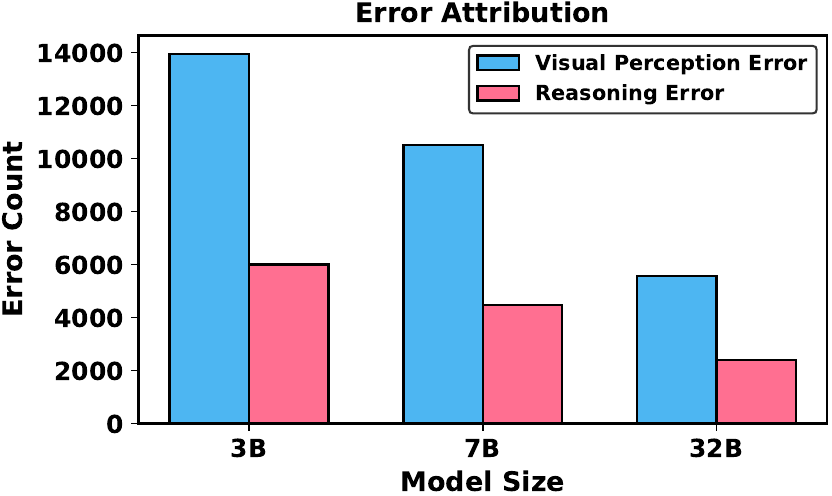}
}
\caption{
GPT-based error attribution of incorrect predictions from Qwen2.5-VL \citep{bai2025qwen2} across different model scales. Results are based on large-scale samples from multiple multimodal reasoning benchmarks, such as ViRL39k \citep{vl-rethinker}, MathV360K \citep{shi2024math}, and Mulberry \citep{yao2024mulberry}, and show that visual perception failures occur \underline{\textbf{more than twice}} as often as reasoning errors.
}
\label{fig:teaser}
\vspace{-0.5cm}
\end{figure}

\section{Introduction}
The reasoning capabilities of Large Vision-Language Models (LVLMs) have advanced considerably through chain-of-thought (CoT) and related techniques that encourage step-by-step problem decomposition \citep{wei2022chain, kojima2022large}. This deliberate analytical approach enables models to solve complex mathematical and scientific problems by breaking them into manageable intermediate steps, and has proven effective across diverse domains, from arithmetic word problems to visually grounded scientific reasoning. However, applying such intensive computation uniformly across all inputs leads to a well-documented issue: overthinking. Models often generate verbose rationales for simple questions that could be answered directly, incurring test-time inefficiency and sometimes introducing errors through excessive elaboration \citep{kahneman2011thinking}.

Overthinking in LVLMs typically manifests in two patterns. First, models may produce redundant explanatory text that restates obvious visual information without advancing toward a solution~\cite{zou2025look}. For instance, when asked to identify a clearly visible object's color, a model may elaborate on the object's shape, position, and context before stating the answer. This behavior is especially prevalent in models trained with extensive CoT data, where the training signal favors detailed explanations regardless of problem complexity. Second, models may engage in unnecessary reasoning chains for problems that primarily require direct visual recognition. A simple counting question can trigger multi-step reasoning about spatial relations and numerical properties when straightforward enumeration would suffice.

Our analysis across multiple challenging multimodal reasoning benchmarks shows that \textbf{a substantial fraction of errors arise from visual perception failures rather than faulty reasoning} (as illustrated in Figure~\ref{fig:teaser}). These failures occur when models misinterpret key visual cues before reasoning begins and are unlikely to be corrected by additional deliberation. This observation motivates our central thesis that adaptive computation in LVLMs should account for perceptual uncertainty alongside reasoning adaptation.

A key challenge in operationalizing this insight is the absence of supervision that distinguishes perceptual failures from reasoning errors. Standard benchmarks provide only final-answer correctness, offering no signal about which cognitive stage failed. To bridge this gap, we derive large-scale failure attribution supervision by mining incorrect predictions from approximately 790k samples drawn from multiple multimodal reasoning benchmarks, such as ViRL39k \citep{vl-rethinker}, MathV360K \citep{shi2024math}, and Mulberry \citep{yao2024mulberry}, and using stronger teacher models to attribute errors to perceptual hallucinations or reasoning failures. This supervision provides targeted signals for learning when additional perceptual or reasoning computation is warranted.

Building on this foundation, we introduce Gated Perception--Reasoning Optimization (GPRO), a framework that extends adaptive reasoning with targeted visual re-analysis. At its core, a meta-reasoning controller evaluates the model’s internal state at each token generation step and routes computation through one of three specialized paths: (1) a lightweight fast path for efficient direct generation; (2) a slow perception path that re-examines visual features to resolve perceptual uncertainty; and (3) a slow reasoning path that engages self-reflection to correct logical errors. This fine-grained control allocates computational resources precisely where they are needed, avoiding unnecessary deliberation while preserving robustness on challenging inputs.

The design of GPRO is inspired by cognitive science, where human problem solving combines fast intuitive responses with slower deliberative reasoning, alongside frequent re-inspection of perceptual inputs under uncertainty \citep{kahneman2011thinking}. By explicitly supporting both perceptual re-analysis and reasoning self-correction, GPRO more closely mirrors this flexible cognitive process. 
Our contributions are summarized as follows:
\begin{itemize}[leftmargin=*]
\item We identify visual perception failures as a critical bottleneck in LVLM performance, largely overlooked by existing adaptive reasoning methods.

\item We derive large-scale failure attribution supervision from approximately 790k samples to distinguish perceptual hallucinations from reasoning errors.

\item We propose the GPRO framework, which dynamically allocates computation between perception and reasoning at token-level granularity via a novel meta-reasoning controller.

\item We demonstrate through extensive experiments on five benchmarks that GPRO models achieve strong accuracy with substantially reduced response lengths, establishing an effective approach for adaptive multimodal reasoning.
\end{itemize}

\section{Related Work}

\subsection{Reasoning in Vision-Language Models}

CoT prompting marked a significant advance in enhancing reasoning capabilities of large models \citep{wei2022chain}. By generating step-by-step rationales before final answers, models can decompose complex problems into intermediate steps amenable to sequential processing. This paradigm has been extended to multimodal settings, where LVLMs generate interleaved text and visual analysis to solve reasoning problems \citep{liu2024visual, bai2025qwen2}. The key insight is that explicit intermediate steps allow models to leverage their language modeling capabilities for multi-hop reasoning that would be difficult in a single forward pass.

Subsequent work has refined this process through various mechanisms. Self-correction approaches \citep{madaan2023selfrefine} enable models to identify and fix errors in their own reasoning paths through iterative refinement \citep{zhou2025variation}. Tree of Thoughts \citep{yao2024tree} explores multiple reasoning paths simultaneously, enabling backtracking and alternative solution exploration when initial approaches fail. These methods have pushed performance boundaries on complex reasoning tasks but often incur substantial computational overhead.

Recent work has advanced reasoning capabilities and post-training adaptation in large language models \citep{comanici2025gemini, diao2025temporal, yuan2025mitigating, yuan2026behavior}. Mulberry \citep{yao2024mulberry} achieves o1-like reasoning through collective Monte Carlo tree search, enabling systematic exploration of solution spaces. Virgo \citep{du2025virgo} explores reproducing slow-thinking mechanisms in multimodal settings through careful data curation and training procedures. LMM-R1 \citep{peng2025lmmr1} employs two-stage rule-based reinforcement learning to enhance reasoning in smaller models. These approaches represent important progress but share a common limitation: they focus primarily on reasoning depth without explicitly addressing perceptual accuracy.

\subsection{Efficiency and Adaptive Computation}

The computational cost of exhaustive reasoning has spurred research on adaptive computation strategies. Mixture-of-Experts architectures \citep{shazeer2017outrageously, fedus2022switch} selectively activate parameter subsets based on input characteristics, enabling capacity scaling without proportional cost increases for every input. The sparse activation pattern provides a natural mechanism for adaptive computation, as different experts can specialize in different input types or reasoning patterns.

Early-exit mechanisms and model cascades represent another approach, routing simpler queries through lightweight processing paths while reserving full model capacity for complex inputs. These methods recognize that not all inputs require the same computational depth and that significant efficiency gains are possible through intelligent routing. Recent work has explored adaptive reasoning strategies specifically for vision-language models. The FAST framework \citep{xiao2025fast} investigates how response length and data distribution affect LVLM performance, developing methods to dynamically adjust reasoning depth based on problem characteristics. Vision-R1 \citep{huang2025visionr1} incentivizes reasoning capability through reinforcement learning on curated multimodal CoT data. Curr-ReFT \citep{deng2025currreft} proposes curriculum reinforcement finetuning to address training bottlenecks in smaller VLMs. Our work extends this direction by explicitly incorporating perceptual uncertainty alongside reasoning adaptation.

\begin{figure*}[t]
  \centering
  \includegraphics[width=\textwidth]{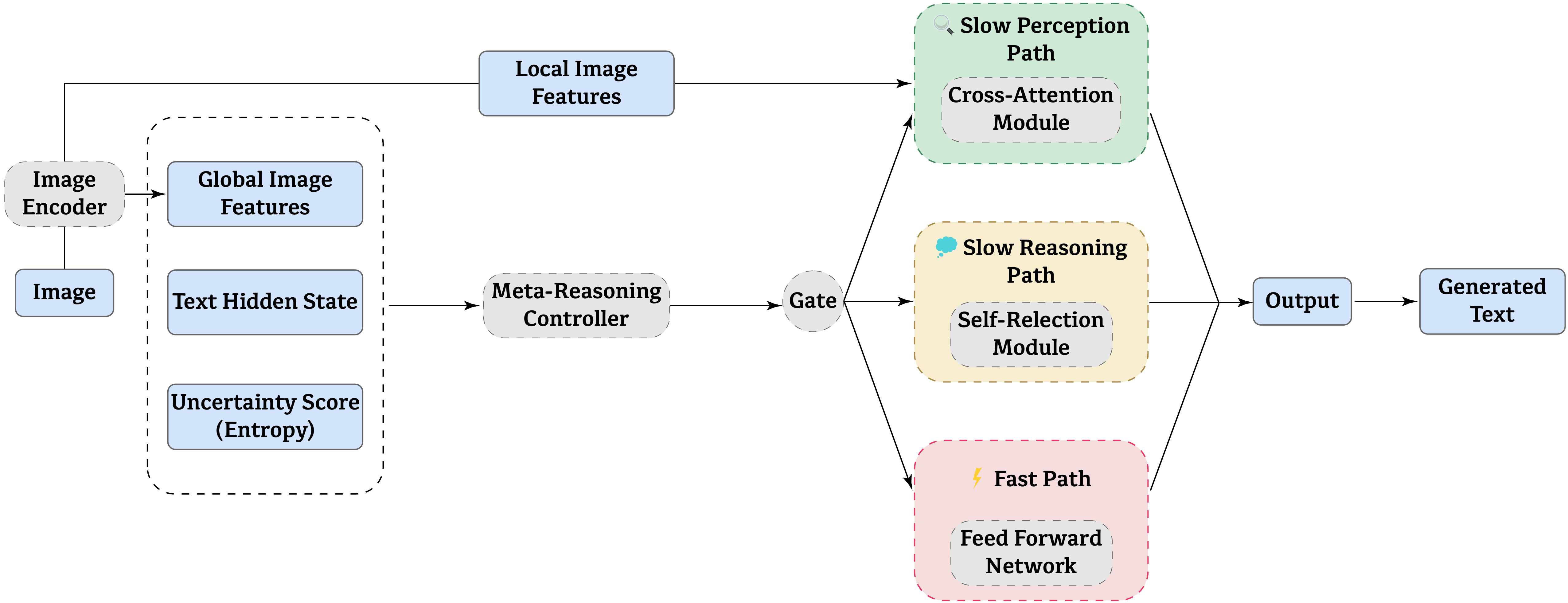}
  \caption{GPRO architecture overview. The meta-reasoning controller receives text hidden states, uncertainty scores, and global image features to route among Fast Path (FFN), Slow Perception Path (Cross-Attention), and Slow Reasoning Path (Self-Reflection).}
  \label{fig:architecture}
\end{figure*}

\subsection{Reinforcement Learning for Language Models}

Reinforcement learning has become central to training adaptive computation frameworks and aligning model behavior with desired objectives \citep{ma2026fipo, hou2025codev, qing2026cluster, diao2025soundmind}. Proximal Policy Optimization \citep{schulman2017proximal} and related algorithms enable models to optimize policies balancing multiple objectives such as accuracy and efficiency. Reinforcement Learning from Human Feedback \citep{ouyang2022training} has proven effective for aligning model behavior with human preferences, demonstrating that RL can shape complex behaviors difficult to specify through supervised learning alone.

Recent multimodal reasoning models have applied RL-based approaches with notable success. R1-OneVision \citep{yang2025r1onevision} advances generalized multimodal reasoning through cross-modal formalization trained with reinforcement learning. MM-Eureka \citep{meng2025mmeureka} explores rule-based reinforcement learning for multimodal reasoning at scale. OpenVLThinker \citep{deng2025openvlthinker} demonstrates that iterative self-improvement through alternating supervised fine-tuning and reinforcement learning can yield sophisticated CoT reasoning capabilities.

Our GPRO framework employs a similar RL-based approach but introduces a more sophisticated state representation incorporating visual perception signals alongside reasoning uncertainty. This enables the controller to trigger dedicated visual re-analysis when perceptual uncertainty is high, a capability absent from prior adaptive reasoning methods that focus solely on reasoning depth.

\section{Method}

We introduce Gated Perception-Reasoning Optimization (GPRO), a framework that augments standard LVLMs with dynamic, fine-grained control over computational resource allocation. The key innovation is a Gated Perception-Reasoning (GPR) module that replaces selected feed-forward layers in the decoder, enabling token-level decisions about whether to invoke additional perception or reasoning computation.

\subsection{Gated Perception-Reasoning Architecture}

The GPR module serves as a lightweight replacement for standard FFN layers, strategically inserted at alternating positions in the Transformer decoder to balance adaptive computation with base model capabilities. This alternating pattern ensures the model maintains its foundational representational power while gaining flexibility to dynamically adjust its computational strategy. Each GPR module comprises a meta-reasoning controller and three computational paths, as illustrated in Figure~\ref{fig:architecture}.

\subsubsection{Meta-Reasoning Controller}

The controller is a compact 2-layer Transformer that determines path selection for each generated token. Its lightweight design ensures minimal overhead while providing sufficient capacity for nuanced decision-making. At timestep $t$, the controller receives a state vector $s_t$ that encapsulates the model's current cognitive state through three complementary signals.

The first signal is the current hidden state $h_t$ from the main decoder, encoding the semantic context of generated text. This captures what the model knows at the current generation step and provides information about the reasoning trajectory so far. The second signal is an uncertainty score $U_t$ computed as predictive entropy over output logits, quantifying the model's confidence in its next token prediction. High entropy indicates the model is uncertain about how to proceed, suggesting additional computation may be beneficial. The third signal is global image features $V_g$ summarizing the visual context, providing the controller with information about the visual input's complexity and content.

Based on the concatenated state $s_t = [h_t; U_t; V_g]$, the controller outputs a discrete action $a_t \in \{\text{fast}, \text{perception}, \text{reasoning}\}$ selecting one of three computational paths.

\subsubsection{Computational Paths}

The three paths address distinct computational needs arising during generation:

\textbf{Fast Path.} The default low-cost option utilizing the original FFN layer from the base model. This path is selected when the model is confident in both its perceptual understanding and reasoning trajectory:
\begin{equation}
\text{Fast}(h_t) = \text{GELU}(h_t W_1 + b_1)W_2 + b_2.
\end{equation}

The fast path preserves the base model's efficient generation capability for straightforward cases.

\textbf{Slow Perception Path.} Activated when the controller detects high perceptual uncertainty, this path uses the current hidden state as a query for cross-attention over visual features:
\begin{equation}
\text{Perc}(h_t, V) = \text{CrossAttn}(h_t, V, V).
\end{equation}

This enables the model to re-examine the image, focusing on details relevant to current generation. The cross-attention mechanism allows targeted retrieval of visual information that may have been overlooked or misinterpreted in initial processing.

\textbf{Slow Reasoning Path.} Engaged when logical uncertainty is high, this path passes the current hidden state and recent context to a meta-reasoning module:
\begin{equation}
\text{Reas}(h_t, H_{<t}) = \text{MetaTrans}(h_t, H_{<t}).
\end{equation}

This supports internal self-reflection, allowing the model to reconsider its reasoning trajectory without generating additional output tokens. The reasoning path can identify and correct logical errors before they propagate to subsequent generation steps.

\subsection{Multi-Objective Training}

The meta-reasoning controller is trained via PPO-based reinforcement learning with a multi-objective reward function designed to balance competing desiderata:
\begin{equation}
R(\tau) = R_{\text{task}} + \alpha_c R_{\text{cost}} + \alpha_l R_{\text{cal}},
\end{equation}
where $\alpha_c$ and $\alpha_l$ are weighting coefficients controlling the trade-off between accuracy and efficiency.

\textbf{Task Reward.} A sparse signal providing the primary learning signal: $R_{\text{task}} = +1$ for correct final answers, $0$ otherwise. This ensures the model prioritizes accuracy as its primary objective.

\textbf{Cost Reward.} Penalizes activation of computationally expensive slow paths:
\begin{equation}
R_{\text{cost}} = -\sum_{t} \left( c_p \mathbb{1}_{[a_t=p]} + c_r \mathbb{1}_{[a_t=r]} \right),
\end{equation}
where $c_p$ and $c_r$ are cost coefficients for perception and reasoning paths respectively. This encourages reliance on the fast path when additional computation is unnecessary.

\textbf{Calibration Reward.} Ensures uncertainty scores reliably indicate when slow paths are needed:
\begin{equation}
R_{\text{cal}} = -\sum_{t \in \mathcal{E}} (1 - U_t) - \sum_{t \in \mathcal{C}} U_t,
\end{equation}
where $\mathcal{E}$ and $\mathcal{C}$ denote tokens leading to incorrect and correct answers respectively. This reward encourages the model to be uncertain when it should be (before errors) and confident when it should be (before correct outputs), enabling the controller to make well-informed decisions.

\subsection{Training Data Construction}

Effective training of the controller requires data that exposes both perceptual and reasoning failure modes. We construct such data through a systematic three-step procedure.

First, we perform failure case mining by running Qwen2.5-VL on a combined dataset of approximately 790K samples drawn from multiple multimodal reasoning benchmarks, such as ViRL39k \citep{vl-rethinker}, MathV360K \citep{shi2024math}, and Mulberry \citep{yao2024mulberry}, collecting all instances where the model produces incorrect answers. This provides a rich set of failure cases spanning diverse visual reasoning challenges. 

Second, we conduct failure attribution using GPT-4 to categorize each failure as either a visual perception failure or a reasoning error propagation. The categorization is based on analysis of the question, image, incorrect answer, and ground truth. Visual perception failures are identified when the model's error can be traced to a misinterpretation of visual elements, while reasoning errors are identified when visual understanding appears correct but the logical steps are flawed. We provide illustrative examples of both failure types in Appendix~\ref{sec:appendix_failure_examples}.

Third, we construct a training curriculum from the labeled data, oversampling difficult examples to ensure the controller encounters sufficient instances of both failure types \citep{jia2025makes}. For the 3B model, we curate approximately 40K training samples, while the 7B model uses approximately 30K samples. The smaller sample size for the larger model reflects its stronger base capabilities requiring less corrective training. 

\begin{table*}[t]
\centering
\small
\setlength{\tabcolsep}{4pt}
\begin{tabular}{l|cc|cc|cc|cc|cc}
\toprule
\multirow{2}{*}{\textbf{Method}} & \multicolumn{2}{c|}{\textbf{MathVision}} & \multicolumn{2}{c|}{\textbf{MathVerse}} & \multicolumn{2}{c|}{\textbf{MathVista}} & \multicolumn{2}{c|}{\textbf{DynaMath}} & \multicolumn{2}{c}{\textbf{MM-Vet}} \\
& Acc. & Len. & Acc. & Len. & Acc. & Len. & Acc. & Len. & Acc. & Len. \\
\midrule
\multicolumn{11}{l}{\cellcolor{gray!15}\textit{Closed-Source Models}} \\
GPT-4o \citep{hurst2024gpt} & 30.4 & -- & \textbf{49.9} & -- & 63.8 & -- & \underline{63.7} & -- & \textbf{80.8} & -- \\
Claude-3.5 Sonnet \citep{claude3} & \underline{37.9} & -- & 46.3 & -- & \underline{67.7} & -- & \textbf{64.8} & -- & 68.7 & -- \\
Qwen-VL-Max \citep{Qwen-VL} & \textbf{39.3} & -- & \underline{47.3} & -- & \textbf{74.2} & -- & -- & -- & \underline{73.2} & -- \\
\midrule
\multicolumn{11}{l}{\cellcolor{orange!15}\textit{Base Qwen2-VL-7B}} \\
Qwen2-VL-7B \citep{wang2024qwen2} & 18.8 & \underline{443.0} & 31.9 & \underline{388.9} & \underline{58.2} & \textbf{265.9} & \underline{39.8} & \underline{298.4} & \textbf{62.0} & \textbf{132.5} \\
Mulberry \citep{yao2024mulberry} & \underline{23.4} & \textbf{349.2} & \textbf{39.5} & \textbf{364.3} & \textbf{62.1} & \underline{275.0} & \textbf{46.8} & \textbf{273.3} & \underline{43.9} & \underline{218.3} \\
Virgo \citep{du2025virgo} & \textbf{24.0} & -- & \underline{36.7} & -- & -- & -- & -- & -- & -- & -- \\
\midrule
\multicolumn{11}{l}{\cellcolor{blue!15}\textit{Base Qwen2.5-VL-3B}} \\
Qwen2.5-VL-3B \citep{bai2025qwen2} & 21.2 & 450.6 & 34.6 & 362.3 & 62.3 & 212.9 & 48.2 & 270.9 & 61.3 & 138.8 \\
Curr-ReFT \citep{deng2025currreft} & 20.1 & \textbf{240.1} & 36.3 & \textbf{121.6} & 61.9 & \textbf{95.9} & 43.8 & \textbf{146.4} & 62.0 & 117.6 \\
LMM-R1 \citep{peng2025lmmr1} & 25.2 & 447.8 & 41.8 & 423.9 & 63.2 & 245.0 & 53.1 & 341.6 & \textbf{65.9} & 166.3 \\
FAST-3B \citep{xiao2025fast} & \underline{26.8} & 323.5 & \underline{43.0} & 286.3 & \underline{66.2} & 158.7 & \underline{54.4} & 213.7 & 64.0 & \underline{112.7} \\
\textbf{GPRO-3B (Ours)} & \textbf{27.1} & \underline{298.6} & \textbf{44.2} & \underline{265.4} & \textbf{66.8} & \underline{145.2} & \textbf{55.1} & \underline{195.3} & \underline{65.2} & \textbf{108.4} \\
\midrule
\multicolumn{11}{l}{\cellcolor{green!15}\textit{Base Qwen2.5-VL-7B}} \\
Qwen2.5-VL-7B \citep{bai2025qwen2} & 25.6 & 443.0 & 46.9 & 388.9 & 68.2 & 189.1 & 58.0 & 273.3 & 67.1 & 132.5 \\
MM-R1 \citep{liang2025mmr1} & 30.2 & 324.6 & \underline{49.8} & 283.9 & 71.0 & 185.6 & 57.5 & 254.2 & 70.6 & 137.9 \\
Vision-R1 \citep{huang2025visionr1} & -- & -- & 52.4 & -- & 73.5 & -- & -- & -- & -- & -- \\
R1-OneVision \citep{yang2025r1onevision} & 29.9 & 692.8 & 46.4 & 631.5 & 64.1 & 402.5 & 53.5 & 560.6 & \textbf{71.6} & 440.7 \\
OpenVLThinker \citep{deng2025openvlthinker} & 29.6 & 457.2 & 47.9 & 398.4 & 70.2 & 305.7 & 57.4 & 382.1 & 68.5 & 312.7 \\
FAST-7B \citep{xiao2025fast} & \underline{30.6} & \underline{204.8} & \textbf{50.6} & \underline{201.0} & \underline{73.8} & \underline{120.7} & \underline{58.3} & \underline{164.8} & \underline{71.2} & \textbf{114.1} \\
\textbf{GPRO-7B (Ours)} & \textbf{31.2} & \textbf{195.6} & 48.7 & \textbf{188.4} & \textbf{74.2} & \textbf{115.3} & \textbf{59.2} & \textbf{158.7} & 70.9 & \underline{118.8} \\
\bottomrule
\end{tabular}
\caption{Main results on five reasoning benchmarks. We report accuracy (\%) and average response length (tokens). Response lengths are measured using Qwen2.5-VL's tokenizer. \textbf{Bold} results indicate the best performance among {all} methods, while \underline{underlined} results indicate the second-best performance among {all} methods.}
\vspace{-0.2cm}
\label{tab:main_results}
\end{table*}

\section{Experiments}

We evaluate GPRO on five challenging benchmarks to address three questions: (1) How does GPRO compare to state-of-the-art methods in accuracy and efficiency? (2) What are the contributions of individual components? (3) How does the model behave qualitatively?

\subsection{Experimental Setup}

\paragraph{Baselines.} We compare against three categories of baselines. Closed-source models include GPT-4o \citep{hurst2024gpt}, Claude-3.5 Sonnet \citep{claude3}, and Qwen-VL-Max \citep{Qwen-VL}, representing current commercial capabilities. Base models include Qwen2-VL-7B \citep{wang2024qwen2} and Qwen2.5-VL \citep{bai2025qwen2} at 3B and 7B scales. Recent slow-thinking methods include Mulberry \citep{yao2024mulberry}, Virgo \citep{du2025virgo}, Curr-ReFT \citep{deng2025currreft}, LMM-R1 \citep{peng2025lmmr1}, MM-R1 \citep{liang2025mmr1}, Vision-R1 \citep{huang2025visionr1}, R1-OneVision \citep{yang2025r1onevision}, OpenVLThinker \citep{deng2025openvlthinker}, and FAST \citep{xiao2025fast}.

\paragraph{Benchmarks.} We evaluate on five benchmarks requiring sophisticated visual and mathematical reasoning: MathVision \citep{wang2024mathvision} for geometric and mathematical problems in visual contexts; MathVerse \citep{zhang2024mathverse} for mathematical reasoning with complex visual diagrams; MathVista \citep{lu2024mathvista} covering diverse mathematical reasoning grounded in visual information; DynaMath \citep{zou2024dynamath} for dynamic mathematical reasoning; and MM-Vet \citep{yu2023mmvet} for integrated multimodal capabilities.

\paragraph{Implementation.} GPRO models are built on Qwen2.5-VL with GPR modules replacing alternate FFN layers. Training used 8 NVIDIA H100 GPUs with batch size 512, 8 rollouts per question, for 10 epochs (approximately 600 GPU hours). Reward weights were $\alpha_c = 0.1$, $\alpha_l = 0.2$, with learning rate $1 \times 10^{-5}$ and cosine decay schedule.

\subsection{Main Results}

Table~\ref{tab:main_results} presents a comprehensive comparison across five multimodal reasoning benchmarks. We report both accuracy (\%) and average response length (tokens) to facilitate a dual analysis of performance and computational efficiency.

The empirical results in Table~\ref{tab:main_results} reveal several critical insights regarding the interplay between reasoning depth and computational efficiency.

\paragraph{Optimizing the Efficiency-Accuracy Frontier.} 
First and foremost, GPRO consistently redefines the Pareto frontier for multimodal reasoning. Unlike prior methods that trade inference speed for accuracy, our approach achieves superior performance while dramatically curtailing token generation. On MathVerse, GPRO-7B improves accuracy by 1.8\% over the base Qwen2.5-VL-7B while reducing the average response length by 51.5\% (from 388.9 to 188.4 tokens). Similarly, on MathVista, we observe a 6.0\% accuracy gain alongside a 39\% reduction in tokens.  This confirms our core hypothesis: model performance is not strictly proportional to generation length, and intelligent resource allocation can mitigate the overthinking phenomenon inherent in standard CoT processes.

\paragraph{Critique of Unconditional Long-Context Reasoning.} 
A comparison with recent long-CoT distillation models, such as R1-OneVision and OpenVLThinker, highlights the limitations of unconditional slow thinking. R1-OneVision exhibits extreme token consumption (e.g., 692.8 tokens on MathVision) yet often underperforms our method (29.9\% vs. 31.2\% for GPRO-7B). This suggests that forcing models to generate extensive reasoning paths for every query introduces noise or hallucination rather than clarity. GPRO's selective activation mechanism acts as a surgical intervention, allocating computational budget only when necessary, thereby achieving a $\sim$3.5$\times$ reduction in inference cost compared to R1-OneVision while surpassing its accuracy.

\paragraph{Competitiveness with Proprietary Giants.} 
Notably, GPRO-7B demonstrates surprising resilience against closed-source models orders of magnitude larger. On MathVision, GPRO-7B (31.2\%) outperforms GPT-4o (30.4\%), and on MathVista, it matches the performance of Qwen-VL-Max (74.2\%). This result is significant as it indicates that smaller, open-weights models, when equipped with efficient meta-reasoning strategies, can bridge the gap with proprietary SOTA models in specific reasoning-heavy domains.

\paragraph{Scalability Across Model Sizes.} 
Finally, the benefits of GPRO are robust across model scales. GPRO-3B outperforms the strong FAST-3B baseline on 4 out of 5 benchmarks and substantially improves upon the base Qwen2.5-VL-3B (e.g., +9.6\% on MathVerse). This scalability suggests that the perception-reasoning decomposition is a fundamental improvement applicable to various architectures, rather than a parameter-scale-dependent optimization. The consistent gains in both the 3B and 7B settings validate the universality of our proposed controller mechanism.

\subsection{Ablation Study}

To deconstruct the efficacy of GPRO and isolate the impact of its constituent modules, we conducted a component-wise ablation study. Table~\ref{tab:ablation} summarizes the performance contributions of the Slow Perception path, Slow Reasoning path, and the Calibration Reward mechanism.

\begin{table}[ht]
\centering
\small
\setlength{\tabcolsep}{6pt}
\begin{tabular}{lcc}
\toprule
\textbf{Configuration} & \textbf{MathVision} & \textbf{MathVerse} \\
\midrule
\textbf{Full GPRO-7B} & \textbf{31.2} & \textbf{48.7} \\
\quad w/o Slow Perception & 27.8 (\textcolor{red}{-3.4}) & 44.3 (\textcolor{red}{-4.4}) \\
\quad w/o Slow Reasoning & 29.5 (\textcolor{red}{-1.7}) & 47.0 (\textcolor{red}{-1.7}) \\
\quad w/o Calibration Reward & 28.9 (\textcolor{red}{-2.3}) & 46.2 (\textcolor{red}{-2.5}) \\
\bottomrule
\end{tabular}
\caption{Ablation study results on MathVision and MathVerse. Numbers in parentheses indicate the absolute accuracy drop compared to the full model configuration.}
\label{tab:ablation}
\vspace{-0.3cm}
\end{table}

\paragraph{Dominance of Visual Grounding.} 
The most significant performance degradation occurs upon removing the \textit{Slow Perception Path}, resulting in a sharp decline of 3.4\% on MathVision and 4.4\% on MathVerse. This finding empirically validates our core hypothesis: visual hallucinations serve as the primary bottleneck in multimodal reasoning. When the model is deprived of the mechanism to re-examine visual features, initial perceptual errors (e.g., misreading axis scales or misidentifying geometric primitives) propagate irreversibly through the reasoning chain. The disparity between the perception drop (-4.4\%) and the reasoning drop (-1.7\%) highlights that current VLMs suffer more from "garbage-in" perceptual failures than from logical deduction errors.

\paragraph{Role of Reflexive Reasoning.} 
The \textit{Slow Reasoning Path}, while less critical than perception, contributes a consistent improvement ($\sim$1.7\% across benchmarks). This module effectively mitigates logic-level errors by enabling self-correction during complex multi-step derivations. However, its lower relative impact suggests that the base Qwen2.5 model possesses robust inherent reasoning capabilities, and the marginal gain from "thinking longer" is diminishing unless grounded in accurate visual data.

\paragraph{Importance of Uncertainty Calibration.} 
Ablating the \textit{Calibration Reward} leads to a substantial performance drop (2.3\%--2.5\%), underscoring the necessity of alignment between the controller's confidence and task difficulty. Without this reward signal, the meta-controller fails to learn an optimal switching policy, often collapsing into a mode of either indiscriminate fast-path usage (under-thinking) or wasteful slow-path activation (over-thinking). The calibration reward effectively regularizes the decision boundary, ensuring that computational overhead is incurred only when model uncertainty is genuinely indicative of potential failure.

\paragraph{Resource Allocation Dynamics.} 
To further understand the controller's behavior, we analyze the path activation distribution on the MathVision test set.  As illustrated in the distribution analysis, GPRO-7B activates the Fast Path for 73\% of tokens, allocating the Slow Perception Path (17\%) and Slow Reasoning Path (10\%) sparsely. This highly skewed distribution confirms that the model has learned a resource-efficient policy, treating slow thinking as a surgical intervention rather than a default mode. Qualitative inspection reveals that Slow Perception is triggered predominantly on high-frequency visual tokens (e.g., numbers in tables, coordinates, legend texts), whereas Slow Reasoning activates during transitional logical connectives (e.g., "therefore", "implies"), demonstrating the semantic awareness of our learned controller.

\subsection{Case Study}

We present two case studies to illustrate how GPRO mitigates overthinking while preserving accuracy through adaptive resource allocation.

\begin{figure}[t]
    \centering
    \includegraphics[width=\columnwidth]{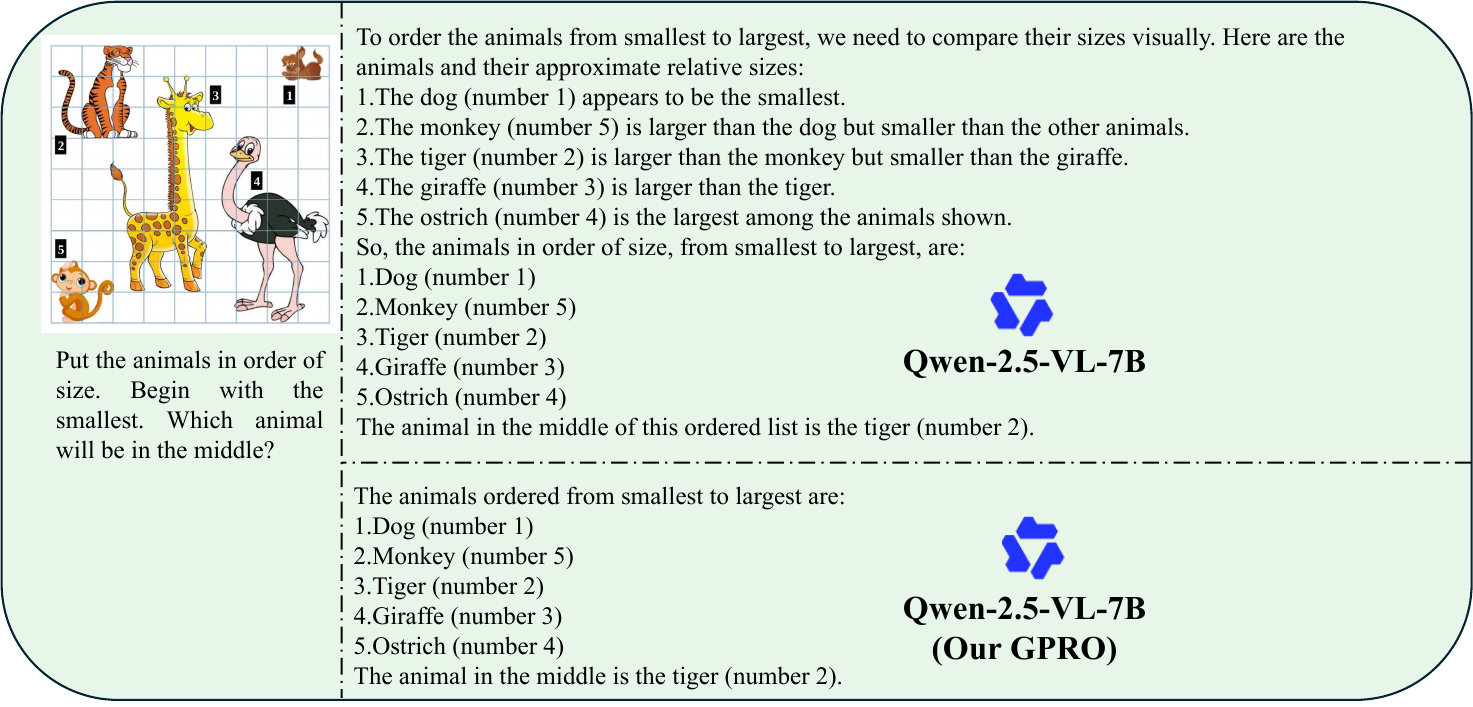}
    \caption{Case Study 1: Animal size ordering. The baseline produces verbose step-by-step comparisons, while GPRO generates a concise direct answer.}
    \label{fig:case_study_1}
    \vspace{-0.35cm}
\end{figure}

\noindent\textbf{Case 1: Visual Ordering Task.}
Figure~\ref{fig:case_study_1} depicts a task that requires ordering animals by size and identifying the middle one.
The baseline Qwen2.5-VL-7B produces lengthy explanations for each animal’s relative size (e.g., ``the dog appears to be the smallest,'' ``the monkey is larger than the dog but smaller than the other animals'') before arriving at the answer.
This behavior exemplifies overthinking, where detailed reasoning is applied to a task that primarily depends on visual perception.
In contrast, GPRO identifies this as a straightforward visual task and predominantly relies on the Fast Path, with selective activation of the Slow Perception Path to verify size relationships when ambiguity arises.
As a result, GPRO produces a correct and concise answer, identifying the tiger as the middle animal while generating substantially fewer tokens.

\begin{figure}[t]
    \centering
    \includegraphics[width=\columnwidth]{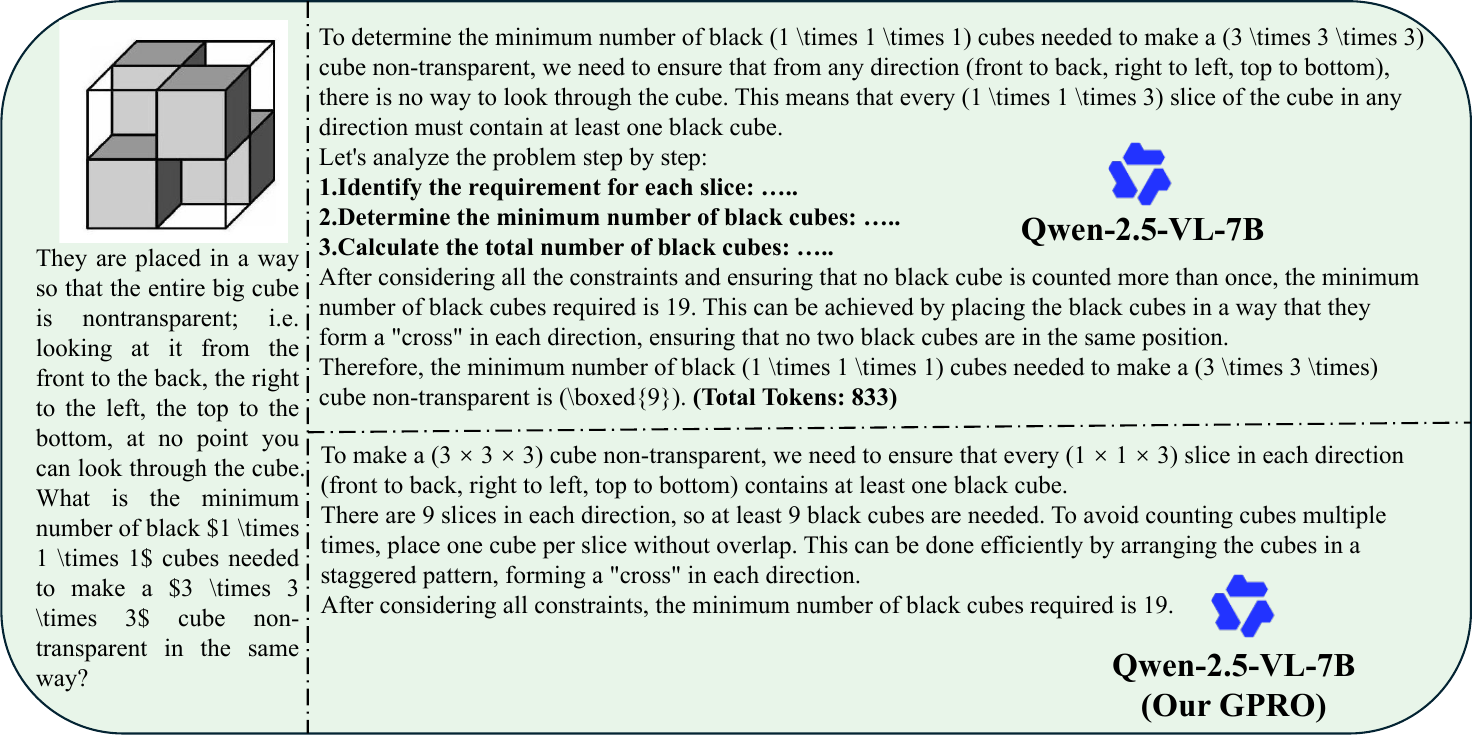}
    \vspace{-0.5cm}
    \caption{Case Study 2: Spatial reasoning with cubes. The baseline generates 833 tokens of detailed analysis, while GPRO produces an efficient solution.}
    \label{fig:case_study_2}
    \vspace{-0.2cm}
\end{figure}

\noindent\textbf{Case 2: Spatial Reasoning.} Figure~\ref{fig:case_study_2} presents a more complex spatial reasoning task that involves counting the number of black cubes required to make a larger cube opaque from all viewing directions. The baseline generates 833 tokens of step-by-step analysis, explicitly enumerating constraints for each slice and direction before reaching a conclusion. GPRO, in contrast, selectively activates the Slow Perception Path to accurately interpret the 3D structure shown in the image, followed by the Slow Reasoning Path to verify the counting logic. The model correctly identifies the key constraint that each slice along every direction must contain at least one black cube, and arrives at the correct answer with significantly fewer tokens. This example illustrates how GPRO coordinates perception and reasoning by invoking additional computation only when needed, leading to a correct solution with substantially reduced generation length.

Together, these cases demonstrate the core advantage of GPRO: dynamically allocating computation based on task demands, avoiding unnecessary verbosity for simple inputs while engaging deeper analysis only when required.

\section{Conclusion}

We introduce Gated Perception--Reasoning Optimization (GPRO), a gated adaptive computation framework for LVLMs.
This work is motivated by the observation that visual perception failures constitute a critical yet underexplored bottleneck in LVLM performance. Such failures often co-occur with overthinking and limit the effectiveness of deeper reasoning alone.
GPRO leverages failure attribution supervision to distinguish perceptual errors from reasoning errors, and uses this signal to guide a meta-reasoning controller that routes token-level generation among fast execution, visual re-examination, and reasoning refinement.
Experiments on five challenging benchmarks show that GPRO improves accuracy while consistently reducing response length, demonstrating an effective approach to adaptive multimodal reasoning.

\section{Limitations}

While GPRO demonstrates strong effectiveness on vision--language reasoning benchmarks, several limitations remain. First, our failure attribution supervision relies on GPT-based labeling, and its quality therefore depends on the reliability of the teacher model's judgments. Although this labeling strategy enables large-scale supervision, GPT-generated attributions may still contain occasional ambiguities, especially in cases where perception and reasoning errors are closely intertwined. Second, GPRO adopts a discrete routing scheme over a fixed set of computation paths. Exploring finer-grained or continuous control over perception and reasoning interventions may offer additional flexibility, which we leave to future work.

\section*{Ethical Considerations}
All experiments presented in this study were conducted using publicly available datasets and models licensed for academic research purposes. To the best of our knowledge, this work does not present any ethical concerns.

\section*{Acknowledgment}
This study is supported by the Department of Defense under Grant No. HT9425-23-1-0267 and in part by the National Science Foundation under Grant No. 2452367.

\bibliography{custom}


\appendix

\section{Benchmarks}
\label{dataset}

\begin{itemize}[leftmargin=*]

    \item {
    \textbf{MathVision} \citep{wang2024mathvision}: MathVision is a multimodal mathematical reasoning benchmark composed of problems collected from real-world math competitions. It spans 16 mathematical disciplines and five difficulty levels, covering diverse visual reasoning scenarios such as geometric constructions, diagrams, graphs, and symbolic annotations.  MathVision requires precise visual grounding followed by multi-step mathematical reasoning, making it particularly sensitive to errors caused by incorrect perception rather than insufficient reasoning.
    }

    \item {
    \textbf{MathVerse} \citep{zhang2024mathverse}: MathVerse is a visual mathematics benchmark that evaluates mathematical reasoning grounded in diagrams. It systematically varies the amount of textual and visual information, forcing models to rely on accurate visual perception rather than textual shortcuts. The benchmark is particularly effective at exposing failures where incorrect visual grounding undermines multi-step reasoning.
    }

    \item {
    \textbf{MathVista} \citep{lu2024mathvista}: 
    MathVista is a benchmark for evaluating mathematical reasoning in visual contexts, comprising problems drawn from 28 existing multimodal datasets and 3 newly curated ones, covering tasks such as geometry, algebra, logic, and scientific reasoning over diverse visual inputs like diagrams, charts, and plots. It is designed to stress fine-grained visual grounding and multi-step mathematical reasoning, where state-of-the-art vision language models still lag behind human performance.
    }

    \item {
    \textbf{DynaMath} \citep{zou2024dynamath}: DynaMath is a dynamic visual mathematics benchmark designed to evaluate the robustness of mathematical reasoning in vision-language models. Each problem is defined as a program that generates multiple concrete variants by altering visual and numerical conditions, such as function parameters, geometric configurations, or graph structures. Unlike prior static benchmarks, DynaMath assesses a model’s ability to generalize across variants of the same underlying problem rather than performance on isolated instances.
    }

    \item {
    \textbf{MM-Vet} \citep{yu2023mmvet}: MM-Vet is a comprehensive multimodal benchmark designed to evaluate the integrated capabilities of large vision-language models. It covers a diverse set of tasks including visual recognition, spatial reasoning, OCR, mathematical reasoning, and commonsense understanding. The benchmark emphasizes multi-faceted reasoning grounded in visual inputs and is widely used for holistic multimodal evaluation.
    }

\end{itemize}

\section{Baselines}
\label{baselines}

\begin{itemize}[leftmargin=*]
    \item {
    \textbf{GPT-4o} \citep{hurst2024gpt}:  
GPT-4o is a large proprietary multimodal foundation model developed by OpenAI, capable of jointly processing visual and textual inputs for a wide range of perception and reasoning tasks. It serves as a strong closed-source baseline for vision--language reasoning, demonstrating robust performance across visual understanding, mathematical reasoning, and multimodal question answering benchmarks.
}

\item {
\textbf{Claude-3.5 Sonnet} \citep{claude3}:  
Claude-3.5 Sonnet is a state-of-the-art proprietary vision--language model developed by Anthropic, designed for high-quality multimodal understanding and reasoning. The model exhibits strong performance on tasks requiring integrated visual perception and multi-step reasoning, and is commonly used as a competitive closed-source baseline in recent multimodal evaluations.
}

\item {
\textbf{Qwen-VL-Max} \citep{Qwen-VL}:  
Qwen-VL-Max is a large-scale proprietary vision--language model from the Qwen family, designed for high-capacity multimodal perception and reasoning. It combines strong visual encoders with advanced cross-modal alignment to support tasks such as visual understanding, OCR, and multimodal question answering. Qwen-VL-Max serves as a competitive closed-source baseline for evaluating multimodal reasoning performance.
}

\item {
\textbf{Qwen2-VL} \citep{wang2024qwen2}:  
Qwen2-VL is an open-weight vision--language model that introduces substantially improved visual encoding and multimodal fusion compared to earlier Qwen-VL models. It enhances visual grounding and cross-modal interaction, enabling stronger performance on a wide range of vision--language understanding and reasoning tasks. Due to its robustness and openness, Qwen2-VL is commonly used as a foundational model in multimodal research.
}

\item {
\textbf{Qwen2.5-VL} \citep{bai2025qwen2}:  
Qwen2.5-VL further advances the Qwen2-VL series by improving visual perception accuracy, instruction following, and reasoning stability. It is trained with enhanced data curation and optimization strategies, resulting in stronger and more consistent performance on challenging multimodal reasoning benchmarks. As a strong open-weight LVLM, Qwen2.5-VL is widely adopted as a base model in recent multimodal reasoning studies.
}

\item {
\textbf{Mulberry} \citep{yao2024mulberry}: Mulberry is a multimodal large language model designed to acquire o1-like reasoning and reflection capabilities. It introduces Collective Monte Carlo Tree Search (CoMCTS), which performs structured search over intermediate reasoning steps by collaboratively expanding and evaluating reasoning paths across multiple models. The resulting reasoning trees are used to construct a large-scale multimodal dataset with explicit intermediate states, enabling supervised training of models with strong multi-step reasoning and self-reflection abilities.
}

\item {
\textbf{Virgo} \citep{du2025virgo}: Virgo is a multimodal slow-thinking model that aims to reproduce o1-like reasoning behavior in vision–language settings through explicit long-form reasoning. It is built by fine-tuning a strong base MLLM (Qwen2-VL) with long chain-of-thought data, primarily text-based reasoning trajectories distilled from existing slow-thinking language models. Rather than introducing new architectural modules or controllers, Virgo relies on instruction tuning to elicit extended step-by-step reasoning during inference. This approach enables Virgo to perform complex multimodal reasoning by generating lengthy natural-language thoughts, but applies slow thinking uniformly across inputs without explicit mechanisms to distinguish perceptual uncertainty from reasoning errors.
}

\item {
\textbf{Curr-ReFT} \citep{deng2025currreft}: Curr-ReFT is a post-training paradigm for improving reasoning and generalization in small- and medium-scale vision-language models via reinforcement learning. It combines Curriculum Reinforcement Learning with difficulty-aware reward schedules, which progressively transition from simple visual judgment to complex open-ended reasoning, and a rejected-sample-based self-improvement stage that preserves core visual and language capabilities. By structuring reinforcement learning along increasing task complexity, Curr-ReFT enables stable optimization and enhances multimodal reasoning without modifying the underlying model architecture.
}

\item {
\textbf{LMM-R1} \citep{peng2025lmmr1}: LMM-R1 is a reasoning-enhancement framework for large multimodal models based on rule-based reinforcement learning. It adopts a two-stage training strategy, first strengthening foundational reasoning using text-only verifiable data, and then generalizing this reasoning capability to multimodal tasks. By decoupling reasoning skill acquisition from multimodal perception, LMM-R1 improves reasoning accuracy in vision-language settings without relying on large-scale high-quality multimodal reasoning annotations.
}

\item {
\textbf{MM-R1} \citep{liang2025mmr1}: MM-R1 is a reasoning-enhanced unified multimodal large language model designed for personalized image generation. It introduces a cross-modal Chain-of-Thought (X-CoT) mechanism that explicitly decomposes generation into an understanding stage and a generation stage, enabling the model to reason about subject identity and attributes before synthesis. MM-R1 further applies reinforcement learning with grouped reward optimization to improve subject fidelity and text-image alignment. Unlike subject-specific fine-tuning methods, MM-R1 supports zero-shot personalization by leveraging intrinsic multimodal reasoning.
}

\item {
\textbf{Vision-R1} \citep{huang2025visionr1}: Vision-R1 is a reinforcement-learning-based multimodal reasoning model that aims to induce chain-of-thought style reasoning in LVLMs. It first performs cold-start initialization using a large automatically constructed multimodal CoT dataset, obtained via modality bridging between an MLLM and a text-only reasoning model. The model is then further optimized with Group Relative Policy Optimization (GRPO), combined with Progressive Thinking Suppression Training to mitigate excessive and incorrect long reasoning chains. Vision-R1 focuses on improving complex multimodal reasoning, particularly in visual mathematical tasks, by encouraging structured and reflective reasoning behavior.
}

\item {
\textbf{R1-OneVision} \citep{yang2025r1onevision}: R1-OneVision is a multimodal reasoning model that explicitly bridges visual perception and language reasoning via a cross-modal formalization pipeline. R1-OneVision converts images into structured textual representations (e.g., formal descriptions of objects, layouts, and relations), enabling precise language-based reasoning over visual content. The model is trained with supervised fine-tuning followed by reinforcement learning to stabilize step-by-step multimodal reasoning and improve generalization across diverse reasoning tasks.
}

\item {
\textbf{MM-R1} \citep{liang2025mmr1}: MM-R1 is a reasoning-enhanced unified MLLM designed for personalized image generation. MM-R1 introduces a cross-modal Chain-of-Thought strategy that explicitly decomposes generation into visual understanding and conditioned synthesis, and further applies GRPO-based reinforcement learning to align subject fidelity and text consistency, enabling zero-shot personalization without subject-specific fine-tuning.
}

\item {
\textbf{OpenVLThinker} \citep{deng2025openvlthinker}: OpenVLThinker is an open-source vision–language model designed to induce \emph{R1-style complex reasoning} in multimodal settings. OpenVLThinker is built on top of Qwen2.5-VL and equips the model with explicit planning, reflection, and verification behaviors by iteratively alternating supervised fine-tuning on chain-of-thought traces and reinforcement learning. Compared to standard LVLMs, it exhibits substantially stronger visual mathematical and logical reasoning while maintaining stable visual grounding.
}

\item {
\textbf{FAST} \citep{xiao2025fast}: FAST proposes a fast–slow thinking framework for LVLMs based on a modified GRPO algorithm. It estimates multimodal question difficulty using both model-dependent signals (pass@k) and image complexity, and applies difficulty-aware length rewards and KL regularization to adaptively control reasoning depth. As a result, the model performs short, efficient reasoning on easy problems while enabling deeper reasoning on hard ones, achieving higher accuracy with significantly reduced response length compared to prior slow-thinking methods.
}

\end{itemize}

\section{Examples of Failure Attribution}
\label{sec:appendix_failure_examples}

\begin{figure*}[t]
  \centering
  \begin{tcolorbox}[
      colback=white,
      colframe=dartgreen,
      title=Visual Perception Failure Example 1,
      boxrule=0.3mm,
      arc=3mm,
      width=\textwidth
    ]

  \centering
  \small\textbf{Image}

  \includegraphics[width=0.6\linewidth]{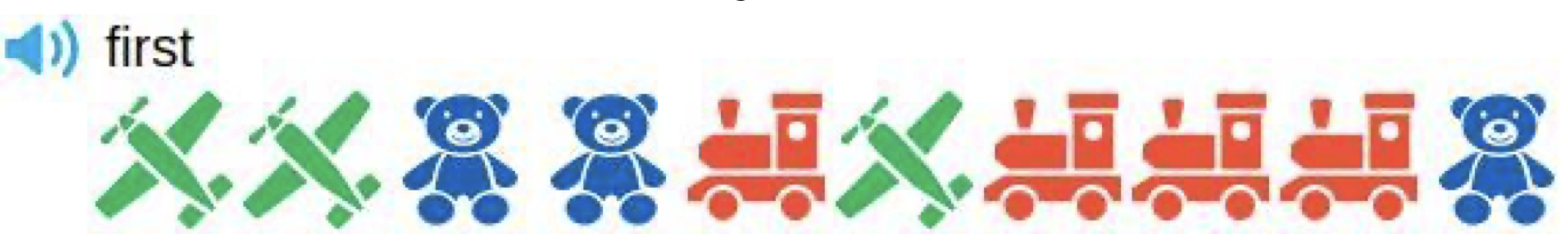}

  \vspace{2ex}
  \noindent\small\textbf{Question}

  \vspace{2.5ex}
  \noindent What is the total number of bears in the image?

  \vspace{3ex}
  \noindent\small\textbf{Anwser by Qwen}

  \small
  \begin{quote}
  To determine the total number of bears in the image, let's count them step by step:

  1. Identify the bears in the image. Bears are typically represented with a rounded shape and sometimes with a face.

  2. Count the bears from left to right:

  • There are two bears at the beginning.

  • There are two more bears in the middle.

  • There is one bear at the end. Adding these up:

  • First set of bears: 2

  • Middle set of bears: 2

  • Last bear: 1

  Total number of bears = 2 + 2 + 1 = 5 So, the total number of bears in the image is 5.
  \end{quote}

  \vspace{1ex}
  \noindent\small\textbf{Error Attribution by GPT}

  \small
  \begin{quote}
  "reasoning": "The model's error is fundamentally rooted in a misinterpretation of the visual data. The image contains a total of 3 bears: two near the beginning (positions 3 and 4) and one at the very end (position 10). The model correctly identifies the first set and the last bear, but it explicitly states there are `two more bears in the middle.' In reality, the objects in the middle are a red train and a green plane. Because the model has hallucinated the existence of bears where there are none (or misidentified other objects as bears), this is a failure of visual grounding rather than a failure of arithmetic logic. The calculation (2+2+1=5) is mathematically consistent with the model's flawed visual premises.", "category": "visual\_perception"
  \end{quote}

  \end{tcolorbox}
  
   \caption{Example of a \textbf{visual perception failure} made by Qwen and the corresponding GPT-based error attribution.}
  \label{fig:visual_perception_error_eg1}
\end{figure*}

\begin{figure*}[t]
  \centering
  \begin{tcolorbox}[
      colback=white,
      colframe=dartgreen,
      title=Visual Perception Failure Example 2,
      boxrule=0.3mm,
      arc=3mm,
      width=\textwidth
    ]

  \centering
  \small\textbf{Image}

  \includegraphics[width=0.15\linewidth]{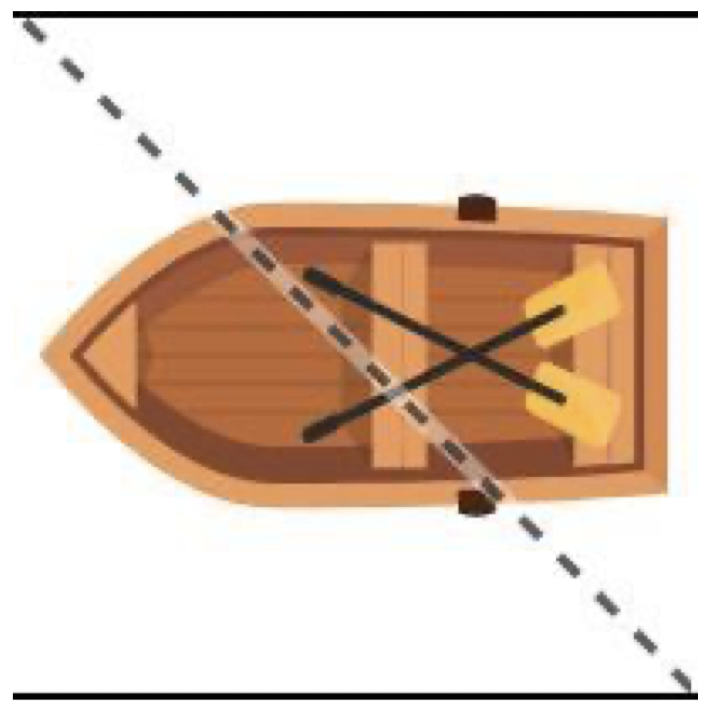}

  \vspace{2ex}
  \noindent\small\textbf{Question}

  \vspace{2.5ex}
  \noindent Are the oars arranged symmetrically in relation to the dotted line?

  \vspace{3ex}
  \noindent\small\textbf{Anwser by Qwen}

  \small
  \begin{quote}
  To determine if the oars are arranged symmetrically in relation to the dotted line, follow these steps:

1. Identify the Symmetry Line: The dotted line in the image is a diagonal line passing through the center of the boat, dividing it into two triangular sections.

2. Examine the Oars: The oars are positioned with their handles at the stern (the back) of the boat and their blades pointing forward. Each oar has a distinct shape and position.

3. Compare the Oars to the Symmetry Line:

• The first oar on the left side of the image is positioned such that its handle is closer to the left side of the boat and its tip is closer to the right side.

• The second oar on the right side of the image is positioned such that its handle is closer to the right side of the boat and its tip is closer to the left side.

4. Evaluate Symmetry:

• For the oars to be symmetrically arranged relative to the dotted line, the oar on the left side should mirror the oar on the right side across the line. The first oar's position is not mirrored by the second oar's position. The first oar's handle is on the left side and the second oar's handle is on the right side, and the first oar's tip is on the right side and the second oar's tip is on the left side.

Given the above analysis, the oars are not arranged symmetrically in relation to the dotted line.
  \end{quote}

  \vspace{1ex}
  \noindent\small\textbf{Error Attribution by GPT}

  \small
  \begin{quote}
"reasoning": "The model's error lies in its visual assessment of the spatial relationship between the oars and the dotted line. The model explicitly claims that `The first oar's position is not mirrored by the second oar's position.' However, visual inspection confirms that the oars are arranged in a crossed formation that is symmetrical with respect to the diagonal dotted line (forming an axis of symmetry). The model fails to correctly perceive this geometric alignment, misinterpreting the orientation and reflection of the objects within the visual space. Since the failure is in recognizing the visual property of symmetry present in the image, rather than a misunderstanding of the concept of symmetry itself, it is a perception failure.", "category": " visual\_perception "
  \end{quote}

  \end{tcolorbox}
  
   \caption{Example of a \textbf{visual perception failure} made by Qwen and the corresponding GPT-based error attribution.}
  \label{fig:visual_perception_error_eg2}
\end{figure*}

\begin{figure*}[t]
  \centering
  \begin{tcolorbox}[
      colback=white,
      colframe=dartgreen,
      title=Reasoning Failure Example 1,
      boxrule=0.3mm,
      arc=3mm,
      width=\textwidth
    ]

  \centering
  \small\textbf{Image}

  \includegraphics[width=0.15\linewidth]{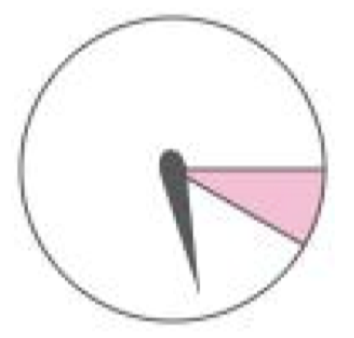}

  \vspace{2ex}
  \noindent\small\textbf{Question}

  \vspace{2.5ex}
  \noindent If the spinner is spun 100 times, how many times can we expect it to land on pink?

  \vspace{3ex}
  \noindent\small\textbf{Anwser by Qwen}

  \small
  \begin{quote}
  To determine how many times we can expect the spinner to land on pink after 100 spins, we need to follow these steps:

1. Identify the probability of landing on pink:

• The spinner is divided into two sections: one pink and one blue.

• The pink section appears to be a smaller portion of the circle compared to the blue section. 

• By visually estimating, the pink section appears to be about 1/4 of the circle.

2. Convert the visual estimate to a probability:

• If the pink section is 1/4 of the circle, the probability ( P ) of landing on pink is ( $\frac{1}{4}$ ).

3. Calculate the expected number of times the spinner will land on pink over 100 spins:

• The expected number of times ( E ) is calculated by multiplying the total number of spins by the probability of landing on pink.

• ( E = 100 $\times$ $\frac{1}{4}$ )

4. Perform the multiplication:

• ( E = 100 $\times$ 0.25 = 25 )

Therefore, we can expect the spinner to land on pink approximately 25 times out of 100 spins.
  \end{quote}

  \vspace{1ex}
  \noindent\small\textbf{Error Attribution by GPT}

  \small
  \begin{quote}
"reasoning": "The model correctly identifies the key visual components: a spinner divided into pink and blue sections. It also correctly retrieves the visual fact that the pink section is smaller than the blue one. However, the error occurs in the logical step of **geometric estimation**. The model arbitrarily assigns a value of `1/4' (90 degrees) to the pink section without applying proper spatial reasoning to verify if the angle is actually a right angle (it is clearly acute, likely around 45-60 degrees). The failure lies not in seeing the color or the sector itself, but in the analytical reasoning process of quantifying that visual area into a probability fraction before performing the final calculation.", "category": "reasoning"
  \end{quote}

  \end{tcolorbox}
  
   \caption{Example of a \textbf{reasoning failure} made by Qwen and the corresponding GPT-based error attribution.}
  \label{fig:reasoning_error_eg1}
\end{figure*}

\begin{figure*}[t]
  \centering
  \begin{tcolorbox}[
      colback=white,
      colframe=dartgreen,
      title=Reasoning Failure Example 2,
      boxrule=0.3mm,
      arc=3mm,
      width=\textwidth
    ]

  \centering
  \small\textbf{Image}

  \includegraphics[width=0.3\linewidth]{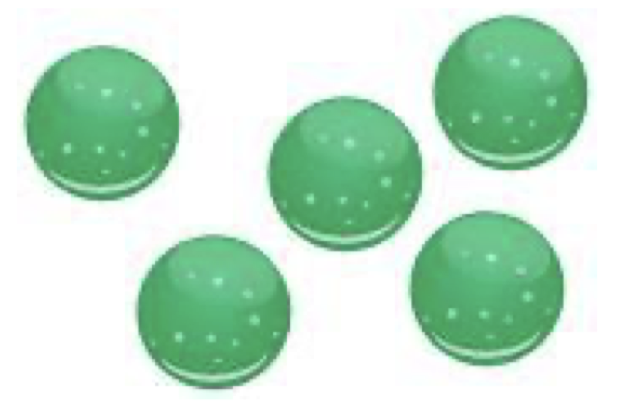}

  \vspace{2ex}
  \noindent\small\textbf{Question}

  \vspace{2.5ex}
  \noindent If three marbles are to be selected without replacement, what is the probability that all three are green?

  \vspace{3ex}
  \noindent\small\textbf{Anwser by Qwen}

  \small
  \begin{quote}
To solve the problem of finding the probability that all three selected marbles are green, we need to follow these steps:

Step 1: Determine the total number of marbles From the image, we can see there are 5 marbles in total.

Step 2: Determine the number of green marbles From the image, we can see that all 5 marbles are green.

Step 3: Calculate the total number of ways to select 3 marbles out of 5 This is a combination problem where the order of selection does not matter. The number of ways to choose 3 marbles out of 5 is given by the combination formula ( $\binom{n}{r}$ = $\frac{n!}{r!(nr)!}$ ), where ( n ) is the total number of items, ( r ) is the number of items to choose, and ( ! ) denotes factorial.

Here, ( n = 5 ) and ( r = 3 ):

$[ \binom{5}{3} = \frac{5!}{3!(5-3)!} = \frac{5!}{3! \cdot 2!} = \frac{5 \times 4 \times 3!}{3! \times 2 \times 1} = \frac{5 \times 4}{2 \times 1} = 10 ]$ So, there are 10 different ways to choose 3 marbles out of 5.

Step 4: Calculate the number of ways to select 3 green marbles out of 5 Since all 5 marbles are green, there is only one way to choose all 3 marbles to be green. This is a straightforward selection since there are no other colors:

[
$\binom{5}{3} = 1$
] 

Step 5: Calculate the probability The probability ( P ) of selecting all 3 green marbles is the ratio of the number of favorable outcomes to the total number of possible outcomes:

$[ P(\text{all 3 green}) = \frac{\text{Number of ways to choose 3 green marbles}}{\text{Total number of ways to choose 3 marbles}} = \frac{1}{10} ]$

Final Answer: The probability that all three selected marbles are green is:

[ $\boxed{\frac{1}{10}}$ ]
  \end{quote}

  \vspace{1ex}
  \noindent\small\textbf{Error Attribution by GPT}

  \small
  \begin{quote}
"reasoning": "The model's visual perception is accurate. It correctly identifies from the image that `there are 5 marbles in total' and 'all 5 marbles are green.' These visual facts are correct. The error occurs purely in the logical reasoning phase, specifically in Step 4. The model incorrectly reasons that if all marbles are green, there is `only one way' to choose 3 green marbles. In reality, if all available marbles are green, *every* possible selection of 3 marbles results in 3 green marbles. Therefore, the number of favorable outcomes (choosing 3 green from 5 green) is identical to the total number of outcomes (choosing 3 from 5), which is 10. The probability should be 10/10 = 1. The model's logic that there is only `one way' to get the result despite correctly identifying the input set size is a flaw in probabilistic reasoning, not visual perception.", "category": "reasoning"
  \end{quote}

  \end{tcolorbox}
  
   \caption{Example of a \textbf{reasoning failure} made by Qwen and the corresponding GPT-based error attribution.}
  \label{fig:reasoning_error_eg2}
\end{figure*}

We provide examples of the two failure types used in our error attribution pipeline. Visual perception failures occur when the model misreads, misidentifies, or incorrectly grounds key visual evidence, as shown in Figures~\ref{fig:visual_perception_error_eg1} and \ref{fig:visual_perception_error_eg2}. In contrast, reasoning errors occur when the model correctly captures the relevant visual information but makes mistakes in the subsequent inference or calculation process, as illustrated in Figures~\ref{fig:reasoning_error_eg1} and \ref{fig:reasoning_error_eg2}. This distinction is important because perception failures call for re-examining the image, whereas reasoning errors require revisiting the intermediate reasoning steps.

\section{Extended Case Study Figures}
\label{sec:appendix_case_study_figures}

\begin{figure*}[ht]
    \centering
    \includegraphics[width=\textwidth]{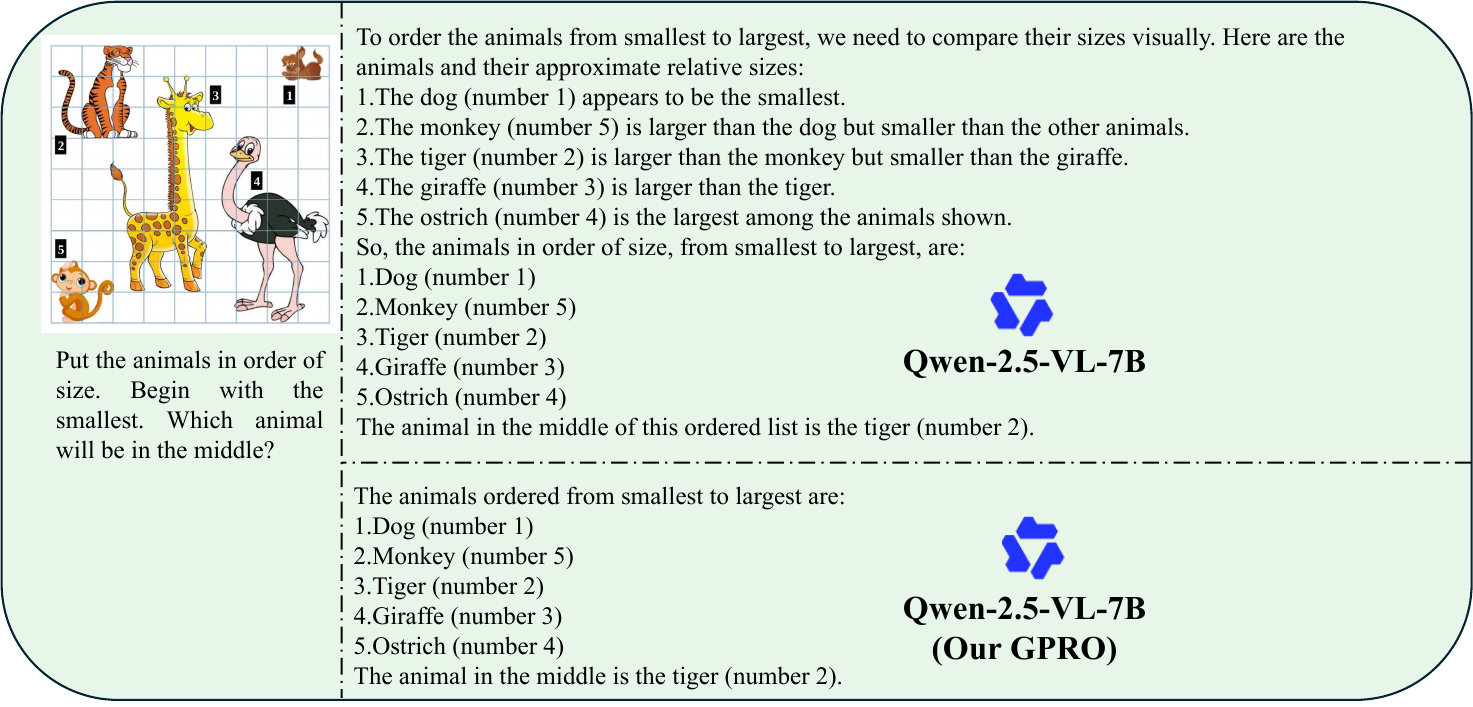}
    \caption{Case Study: Animal size ordering. The baseline produces verbose step-by-step comparisons, while GPRO generates a concise direct answer.}
    \label{fig:case_study_1_big}
\end{figure*}

\begin{figure*}[ht]
    \centering
    \includegraphics[width=\textwidth]{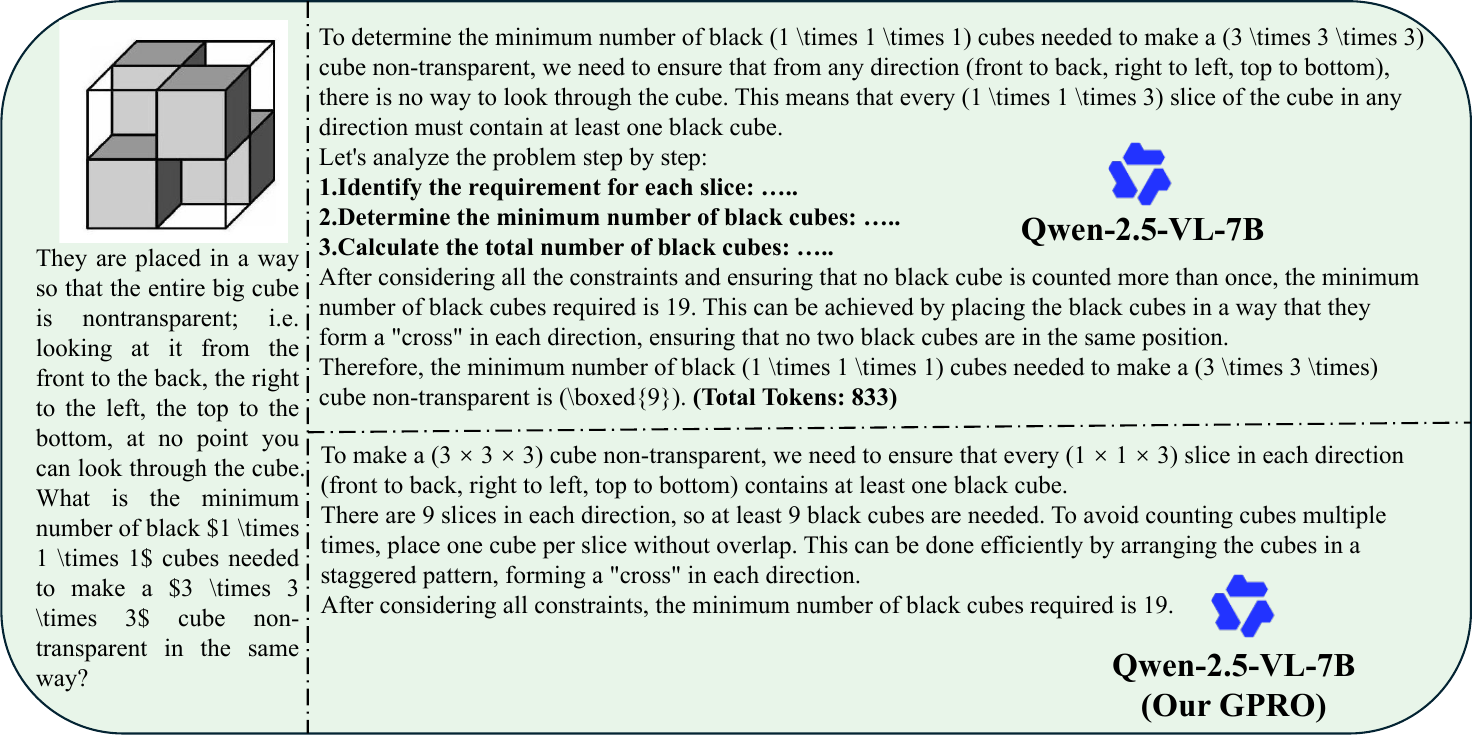}
    \caption{Case Study: Spatial reasoning with cubes. The baseline generates 833 tokens of detailed analysis, while GPRO produces an efficient solution.}
    \label{fig:case_study_2_big}
\end{figure*}

We include enlarged versions of the case study figures from the main paper in Figures~\ref{fig:case_study_1_big} and \ref{fig:case_study_2_big} for improved readability and closer inspection of the visual details and model outputs.

\clearpage

\clearpage

\clearpage

\clearpage

\end{document}